%% file: acl_latex.tex
\pdfoutput=1
\PassOptionsToPackage{x11names}{xcolor} % modsf

\documentclass[11pt]{article}

\usepackage[]{acl}

\usepackage{times}
\usepackage{latexsym}

\usepackage[T1]{fontenc}
\usepackage[utf8]{inputenc}

\usepackage{microtype}

\usepackage{pifont}
\usepackage{hyperref}
\usepackage{bm}
\usepackage{multirow}
\usepackage{graphicx}
\usepackage{subcaption}
\usepackage{float}
\usepackage{quoting}
\usepackage{tcolorbox}
\usepackage[x11names]{xcolor}
\usepackage{booktabs}      % borders in tabular

\definecolor{mgray}{RGB}{192,192,192}

\newcommand{\LN}{\linebreak\noindent}    % manage inline spacing
\usepackage[hang,flushmargin]{footmisc}  % minimize footnote indentation

\newcommand{\hum}[1]{\textcolor{Blue3}{\textbf{\texttt{(S1)}} #1}}
\newcommand{\sys}[1]{\textcolor{black!70!green}{\textbf{\texttt{(S2)}} #1}}
\newcommand{\res}[1]{\textcolor{black}{\textbf{\texttt{[R]}} #1}}

\usepackage{enumitem}      % list margins
\setenumerate[1]{leftmargin=*}    % no enumerate indentation
\setitemize[1]{leftmargin=*}      % no itemize indentation
\usepackage{colortbl}

\renewenvironment{quote}
  {\small\list{}{\rightmargin=-1em \leftmargin=-1em}%
   \item\relax}
  {\endlist}

\title{Leveraging Large Language Models for Automated Dialogue Analysis}

\author{Sarah E. Finch \hspace{2em} Ellie S. Paek \hspace{2em} Jinho D. Choi \\
  Department of Computer Science \\
  Emory University \\
  Atlanta, GA, USA \\
  \texttt{\{sfillwo, ellie.paek, jinho.choi\}@emory.edu}
  }

\begin{document}
\maketitle

\input{latex/abstract}

\input{latex/introduction}
\input{latex/related_work}

\input{latex/dataset}
\input{latex/models}
\input{latex/evaluation}

\input{latex/error_analysis}
\input{latex/recommendation}
\input{latex/limitations}
\input{latex/conclusion}
\input{latex/acknowledgements}

\bibliography{custom}
\bibliographystyle{acl_natbib}

\cleardoublepage
\appendix
\input{latex/appendix}

\end{document}

%% file: latex/abstract.tex
\begin{abstract}

Developing high-performing dialogue systems benefits from the automatic identification of undesirable behaviors in system responses. 
However, detecting such behaviors remains challenging, as it draws on a breadth of general knowledge and understanding of conversational practices.
Although recent research has focused on building specialized classifiers for detecting specific dialogue behaviors, the behavior coverage is still incomplete and there is a lack of testing on real-world human-bot interactions.
This paper investigates the ability of a state-of-the-art large language model (LLM), \LN ChatGPT-3.5, to perform dialogue behavior detection for nine categories in real human-bot dialogues.
We aim to assess whether ChatGPT can match specialized models and approximate human performance, thereby reducing the cost of behavior detection tasks.
Our findings reveal that neither specialized models nor ChatGPT have yet achieved satisfactory results for this task, falling short of human performance.
Nevertheless, ChatGPT shows promising potential and often outperforms specialized detection models.
We conclude with an in-depth examination of the prevalent shortcomings of ChatGPT, offering guidance for future research to enhance LLM capabilities.

%Developing dialogue systems with high performance requires the ability to automatically identify undesirable behaviors in system responses.
%However, detecting such behaviors in conversational AI remains a challenging task, as it draws on a breadth of general knowledge and understanding of conversational practices.
%Recent research has focused on developing specialized classifiers for detecting specific dialogue behaviors, but there is a lack of coverage on important dialogue behaviors and on bot responses from realistic deployment settings.
%In this paper, we investigate the ability of large language models (LLMs), such as GPT3, to perform dialogue behavior detection for 9 behaviors on real human-bot dialogues.
%We aim to explore whether LLMs can perform as well as specialized models and approximate human-level performance, thus reducing the overall cost of behavior detection tasks.
%Our results show that both specialized models and LLMs for dialogue behavior detection have not yet achieved satisfactory results, falling short of human performance.
%However, LLMs show promising potential and often surpass the performance of specialized detection models.
%We conclude with an in-depth study of the prevalent shortcomings of LLMs, which can guide future work towards improving their capability.

\end{abstract}

%% file: latex/introduction.tex
\section{Introduction}

% being able to identify errors in dialogue is useful (development, evaluation, etc)
% certain behaviors get a lot of attention (fact-checking) vs others (which have neither labelled datasets nor detection models)
% many dialogue behaviors / errors are more subjective than other NLP classification tasks (irrelevant, contradiction, commonsense, empathetic, ...) which can lead to lack of objective "ground-truth" -- rather aggregates over many human annotators represents the general human consensus
% can a detection model approximate a human?
% can pretrained large langauge models approximate human?
% compare various models to these human annotators in aggregate
% abc-eval reports moderate to high agreement between humans on all behaviors, and provides multiple human annotations for a subset of the dialogues

%%%%%%%%%%%%%%%%%%%%%%%%%%%%%%%%%%%%%%%%%%%%%%%%%%%%%%%%%%%%%%%%%%%%%%%%%%%%%%%%%%%%%%%%%%%%%%%%%%%%

\begin{figure}[hpt!]
    \centering
    \includegraphics[width=\columnwidth]{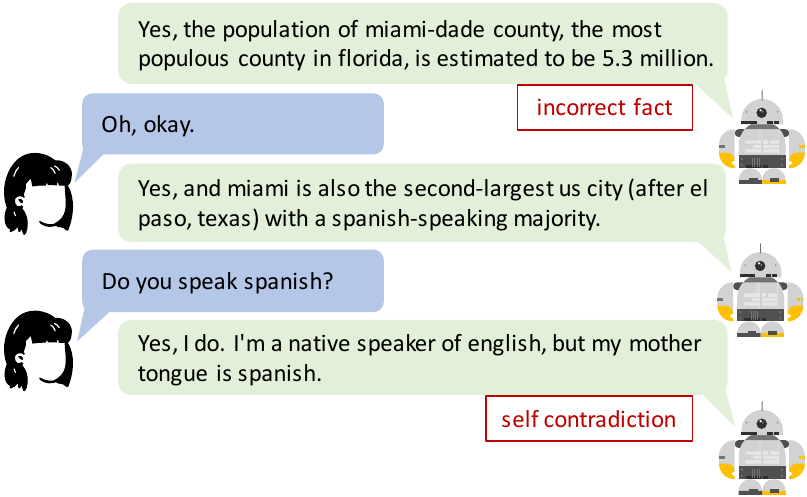}
    \caption{Response errors in a human-bot dialogue.}
    \label{fig:example_dialogue}
    % \vspace{-1.5em}
\end{figure}

One crucial aspect of developing high-performing dialogue systems is the automated identification of errors in system responses. 
These errors can result from various behaviors, including incorrect information retrieval or illogical semantics (Figure~\ref{fig:example_dialogue}).
Identifying such errors enhances dialogue system development and complements dialogue-level evaluation methods by providing finer-grained metrics for comparison \cite{finch:23}.

%A crucial aspect of developing high-performing dialogue systems is the ability to automatically identify errors in the system's responses.
%Errors can arise due to a variety of behaviors such as incorrect information retrieval or misunderstanding the user's intent. 
%Identifying these errors is not only useful for the development of dialogue systems but also complements the standard dialogue-level Likert-score or pairwise selection evaluations for dialogue systems by providing finegrained metrics on which to compare \cite{finch:23}.

%%%%%%%%%%%%%%%%%%%%%%%%%%%%%%%%%%%%%%%%%%%%%%%%%%%%%%%%%%%%%%%%%%%%%%%%%%%%%%%%%%%%%%%%%%%%%%%%%%%%

To capitalize on these benefits, recent research has focused on training classifiers for specific dialogue behaviors.
While certain behaviors have received considerable attention, this is not the case for all pertinent dialogue behaviors. 
Furthermore, most datasets for training are produced by annotating human-human dialogues \cite{sharma:20}, perturbing human responses \cite{gupta:22}, or crafting post-hoc responses \cite{nie:21}. As a result, such datasets may not reflect human-bot interactions, rendering them less suitable for classifier development.

%To address these challenges, recent research has focused on developing classifiers that can approximate human performance in identifying specific dialogue behaviors.
%While certain behaviors have received significant attention, others are currently underexplored, resulting in a lack of coverage for automated detection of dialogue behaviors. 
%In addition, most existing datasets are produced by either (1) annotating human-human dialogues \cite{sharma:20}, (2) perturbing human responses \cite{gupta:22}, or (3) writing post-hoc responses \cite{nie:21}. 
%Such datasets are not necessarily representative of human-bot dialogues in the wild, thus rendering them suboptimal benchmarks for such models. 

%%%%%%%%%%%%%%%%%%%%%%%%%%%%%%%%%%%%%%%%%%%%%%%%%%%%%%%%%%%%%%%%%%%%%%%%%%%%%%%%%%%%%%%%%%%%%%%%%%%%

Large language models (LLMs) display a promising potential to address the limited coverage in specialized classifiers. 
LLMs have demonstrated competitive performance across various natural language processing (NLP) tasks without finetuning \cite{kocon:23}. 
%It is plausible that LLMs encode the necessary knowledge for accurately identifying dialogue behaviors in human-bot dialogues. 
Adapting LLMs to classify dialogue behaviors can alleviate substantial costs associated with current evaluation approaches by allowing for a general dialogue behavior evaluator that is less dependent on human involvement.

Although there is much effort towards open-sourcing competitive LLMs, OpenAI's ChatGPT remains the most successful LLM to date \cite{wang:23}. Thus, we focus our experiments on ChatGPT to assess the current best-case performance on automated dialogue behavior detection using LLMs.
% This paper explores the effectiveness of multiple models in identifying various dialogue behaviors.
With its wide accessibility and low costs, ChatGPT provides a practical and straightforward platform for automating dialogue behavior detection, if its proves successful.

\noindent To this end, our work focuses on two main objectives:

\begin{enumerate}
\setlength\itemsep{0em}
\item To determine whether or not ChatGPT can match the performance of state-of-the-art specialized behavior classifiers.
\item To assess the extent to which ChatGPT can approximate human-level performance in behavior classification using real human-bot dialogues.
\end{enumerate}

\noindent Our findings indicate that automated methods for dialogue behavior detection have not reached satisfactory results, falling short of human performance. 
However, ChatGPT showcases compelling results comparative to or often better than specialized models.
To facilitate further advancements, we conduct an in-depth analysis to identify the prevalent errors and shortcomings of ChatGPT.
This analysis provides valuable insights, highlighting key areas to be targeted to enhance the performance of LLMs in dialogue behavior detection for future work.   We release our code and data at \url{https://github.com/emorynlp/GPT-ABCEval}.

%% file: latex/related_work.tex
\begin{table*}[ht!]
\resizebox{\textwidth}{!}{%
\begin{tabular}{l|r|l}
\toprule
\multicolumn{1}{c|}{\textbf{Label}} & \multicolumn{1}{c|}{\textbf{Abbr.}} & \multicolumn{1}{c}{\textbf{Description}} \\
\midrule
\bf Empathetic & \tt Emp & The response shows an understanding and reacts appropriately to someone's emotions. \\ 
\bf Lack of Empathy & \tt !Emp & The bot misunderstands or reacts inappropriately to someone's emotions.  \\
\bf Commonsense & \multirow{2}{*}{\tt !Com} & \multirow{2}{*}{The response misunderstands or contradicts common knowledge.}  \\ 
\bf Contradiction & & \\ 
\bf Incorrect Fact & \tt !Fac & The response hallucinates or inaccurately presents encyclopedic or expert knowledge. \\ 
\bf Self Contradiction & \tt !Sel & The bot contradicts something it said earlier in the dialogue. \\
\bf Partner Contradiction & \tt !Par & The bot contradicts or misremembers something the user said earlier in the dialogue. \\
\bf Redundant & \tt Red & The response inappropriately repeats information presented earlier in the dialogue. \\
\bf Ignore & \tt Ign & The response ignores what the user just said. \\
\bf Irrelevant & \tt !Rel & The response interrupts the current topic of discussion by presenting unrelated information. \\
\bottomrule
\end{tabular}
}
\caption{The 9 behavior labels from ABC-Eval (table adapted from \citet{finch:23}). The \{\texttt{Emp}, \texttt{!Emp}\}, \{\texttt{!Fac}\}, \{\texttt{!Sel}\}, \{\texttt{Ign}, \texttt{!Rel}\} labels can be classified by the EPI, FC, DEC, S2T2 models in Section~\ref{sec:models}, respectively.}
\label{tab:behavior_labels}
\vspace{-1em}
\end{table*}

\section{Related Work}
\label{sec:rw}

ChatGPT has shown promising performance on many NLP tasks, especially for text classification \cite{gilardi:23, kocon:23, zhu:23}.
In addition, GPT models, including ChatGPT and InstructGPT, have been used to produce high-quality dyadic dialogues \cite{kim:22, zhan:23} and have been shown to correlate highly with human annotators when evaluating the overall quality of empathetic dialogues \cite{svikhnushina:23}. 
However, ChatGPT still exhibits limitations as \citet{chan:23} show that ChatGPT struggles with fine-grained dialogue understanding, reporting poor performance on classifying discourse structure and utterance relations. 

%GPT has shown promising performance on a variety of NLP tasks, especially for text classification tasks \cite{gilardi:23, kocon:23, zhu:23}. 
%% although it has been seen that pragmatic-leaning tasks such as emotion-related classifications do not produce as compelling of results
%For dialogue in particular, ChatGPT has been shown to produce high-quality dyadic dialogues \cite{kim:22, zhan:23} and to achieve compelling system-level correlations with human annotators when classifying the overall quality of empathetic dialogues \cite{svikhnushina:23}. 
%On the other hand, \citet{chan:23} show that ChatGPT struggles with fine-grained dialogue understanding, reporting poor performance on classifying discourse structure and utterance relations. 

To the best of our knowledge, no prior research has explored the use of any GPT model as a behavior classifier for chatbot responses. 
Instead, previous work has focused on the development of specialized dialogue behavior classifiers, as discussed in this section.

%To the best of our knowledge, no work has sought to use ChatGPT as a behavior classifier for chatbot responses. 
%More commonly, previous work has studied the development of specialized dialogue behavior classifiers, as discussed next.  

\subsection{Contradiction Detection}
\label{sec:rw_speaker_contradiction}

Although much work focuses on dialogue contradictions in the context of a given bot persona \cite{zhang:18, welleck:19, kim:20, song:20, shuster:22}, there has been some work on a more general sense of contradictions, including NLI models targeting self-context contradictions \cite{li:21, nie:21}, inconsistency detectors using domain-specific attribute-value classifiers \cite{shi:21}, and context summarization to encourage consistency in response generation \cite{xu_beyond:22, xu_long:22}. Notably, these existing approaches to contradiction detection fail to address partner contradictions.

There is also a lack of work on general commonsense contradiction detection for dialogue responses. To the best of our knowledge, \citet{ghazarian:23} is the only work that focuses explicitly on capturing commonsense qualities of dialogue responses. They propose a method for calculating continuous event commonsense alignment scores for dialogue responses using similarity calculations with the outputs of an event extraction model and generative commonsense model. However, such continuous scores cannot be immediately applied to commonsense contradiction detection without further modifications (e.g. learned thresholding, classification head, etc.).

% fine-grained contradictions categories of intra-sentence contradiction, role confusion, and history for Chinese dialogues \cite{zheng:22}.

% Approaches to natural language entailment are the most closely related to commonsense contradictions, but the dialogue-specific entailment datasets are restricted to only persona- and self-entailment relations \cite{welleck:19, nie:21}. 

\subsection{Claim Verification}
\label{sec:rw_claim_verification}

There are a variety of approaches taken for claim verification in dialogue, including question-answering \cite{honovich:21} and trained classifiers \cite{dziri_begin:22}. \citet{dziri_begin:22} find that trained classifiers perform the best, although they still lag behind human performance. Some works focus on claim verification for question-response pairs only \cite{wang:22}, whereas others target multi-turn dialogues, producing annotated datasets including FaithDial \cite{dziri_faithdial:22}, BEGIN \cite{dziri_begin:22}, and DialFact \cite{gupta:22}. Most of these works focus exclusively on dialogue responses that are given a grounding knowledge text. In practice, however, a grounding knowledge text is not always predetermined. \citet{gupta:22} propose a pipeline for claim verification that includes a knowledge retrieval stage rather than assuming it is provided.

\subsection{Empathy}
\label{sec:rw_empathy}

Human judges are commonly used when evaluating the degree of empathy exhibited in a dialogue response \cite{zhong:20,sabour:22,qian:23}. There has also been some work on developing empathetic response and question taxonomies, although these are only applied in small-scale or synthetic settings \cite{welivita:20, svikhnushina:22}. Most applicably, \citet{sharma:20} collect EPITOME, a dataset of 10K interactions from Reddit and Talklife (a mental health forum) that are annotated with the strength of their expression of three empathetic mechanisms: reactions, interpretations, explorations. Some recent dialogue works have used EPITOME-trained classifiers in their approaches \cite{zheng:21, majumder:22} or for automatic evaluation \cite{kim:21, lee:22}.

\subsection{Coherence}
\label{sec:rw_coherence}

Research on detecting incoherent behaviors, such as redundancy and irrelevancy, is limited. Most works perturb dialogue responses to artificially construct incoherence examples \cite{xu:21, zhang:21, ghazarian:22}, which may not produce representative examples. On the other hand, \citet{mehri:20} derive a response's relevancy score from the probabilities of manually designed future indicator utterances but found little correlation with human judgments. In addition, detection of response redundancy is underexplored, despite some works addressing token repetition \cite{li:20, xi:21}. Perhaps most relevant, the Dialogue Breakdown Detection Challenge (DBDC) aims to identify contextually inappropriate bot responses that hinder conversation continuation \cite{higashinaka:19}. Various classifiers have been proposed for this challenge \cite{ng:20,lin:22}, with observations suggesting coherence issues as a dominant cause of breakdowns.

%% file: latex/dataset.tex
\section{ABC-Eval Dataset}
\label{sec:dataset} 

We use the ABC-Eval dataset from \citet{finch:23} as the behavior detection benchmark. This dataset contains 400 open-domain human-bot dialogues collected between university students and one of four chatbots: BlenderBot2, Blenderbot using DECODE reranking, Emora, and Bart-FiD-RAG. For each bot response in each dialogue, human annotators labeled whether or not a specific dialogue behavior was present. These turn-level binary annotations were collected using crowdworking annotators on the SurgeHQ platform,\footnote{\url{https://www.surgehq.ai}} who were trained on three curated conversations to accurately identify each dialogue behavior before being accepted into the annotation project. For example, in Figure \ref{fig:example_dialogue}, the three bot responses are labeled \texttt{1}, \texttt{0}, \texttt{0} for the behavior \texttt{incorrect fact (!Fac)} and are labeled \texttt{0}, \texttt{0}, \texttt{1} for the behavior  \texttt{self contradiction (!Sel)}.

In this work, we take 1,634 bot responses from 108 dialogues that received two rounds of human annotations, and focus on the nine dialogue behaviors that \citet{finch:23} found as the most informative for capturing dialogue quality (Table~\ref{tab:behavior_labels}).

%We utilize ABC-Eval from \citet{finch:23} in this work as behavior detection benchmark. ABC-Eval is a dataset of 400 open-domain human-bot conversations, collected between university students and one of 4 chatbots: BlenderBot2, Blenderbot using DECODE reranking, Emora, and Bart-FiD-RAG. It includes turn-level human-annotations on 15 bot responses per dialogue for many behaviors, collected using annotators from the annotation company SurgeHQ who were trained on a curated set of conversations to accurately identify each dialogue behavior before being accepted into the annotation project. In this work, we utilize the 1634 chatbot responses from 100 dialogues that received two rounds of human annotations and we focus on the 9 dialogue behaviors that were found to be the most informative towards capturing overall dialogue quality in Table \ref{tab:behavior_labels}.

% To the best of our knowledge, two datasets exist that annotate several turn-level characteristics of human-bot dialogues: the FED dataset \citet{mehri:20} and the ABC-Eval dataset \cite{finch:23}. The turn-level annotations for FED do not label dialogue behaviors, however; instead, they label the semantic characteristics (e.g. interestingness, genericness, etc.). On the other hand, 

%% file: latex/models.tex
\section{Specialized Behavior Detection Models}
\label{sec:models}

% \begin{table}[h!]
%     \centering
%     \begin{tabular}{r|ll}
%     \bf Error & \bf Model & \\ \hline \hline
%     \tt !Fac & \underline{Faith-Critic} & (\S\ref{sec:rw_claim_verification}) \\
%     \tt Red & n/a & (\S\ref{sec:rw_coherence})\\
%     \tt Ign & \underline{S2T2} & (\S\ref{sec:rw_coherence}) \\
%     \tt !Com & n/a & (\S\ref{sec:rw_commonsense_contradiction}) \\
%     \tt !Rel & \underline{S2T2} & (\S\ref{sec:rw_coherence})\\
%     \tt !Par & n/a & (\S\ref{sec:rw_speaker_contradiction})\\
%     \tt !Sel & \underline{DECODE} & (\S\ref{sec:rw_speaker_contradiction})\\
%     \tt Emp & EPITOME & (\S\ref{sec:rw_empathy})  \\
%     \tt !Emp & EPITOME & (\S\ref{sec:rw_empathy}) \\
%     \end{tabular}
%     \caption{State-of-the-art classification models matched to the dialogue behavior types that they most closely target. \underline{Underline} denotes trained models that were released; otherwise, models were trained using the available code. Tasks for which there is no similar previous work are indicated with \textit{n/a}. }
%     \label{tab:sota_models}
% \end{table}

% Table \ref{tab:sota_models} indicates the state-of-the-art models for detecting the dialogue behaviors included in this work. 

In this section, we present state-of-the-art models designed to classify labels that closely align with six of the dialogue behaviors in Table~\ref{tab:behavior_labels}: \texttt{Emp}, \texttt{!Emp}, \texttt{!Fac}, \texttt{!Sel}, \texttt{Ign}, and \texttt{!Rel}.
Note that no existing models are available for predicting\LN \texttt{!Com}, \texttt{!Par}, and \texttt{Red} so there are no viable comparisons to our LLM approach for them (Section~\ref{sec:gpt_llm}).

%Next,we provide details on the state-of-the-art classification models matched to the dialogue behaviors that they most closely target. 
% It is important to note that not all dialogue behaviors have been adequately explored in the relevant literature, thus leading to gaps in the coverage of classifiers for \texttt{!Par}, \texttt{!Com}, and \texttt{Red}.

\paragraph{FaithCritic (FC)} 

Following \citet{gupta:22}, we build a claim verification pipeline for a dialogue response $r$.
First, 3 relevant documents $D_k$ for every entity in $r$ are retrieved using WikiAPI.
Then, a BERT model trained on the Wizard of Wikipedia (WoW) knowledge-response pairs \cite{dinan:19} selects the top-10 evidence sentences $S_e$ from $D_k$.
To distinguish whether a response makes a factual claim or not, the lexical overlap between $r$ and $S_e$ is estimated, optimized on the ABC-Eval training conversations.
Finally, a RoBERTa model trained on Faith-Critic, a dataset of human-annotated faithful and unfaithful evidence-response pairs derived from the WoW \cite{dziri_faithdial:22}, is applied to those responses that make factual claims. 
As a result, responses that are predicted unfaithful to any evidence $e \in S_e$ are labeled as \texttt{!Fac}.

%We construct a claim verification pipeline for a given dialogue response $r$ following \citet{gupta:22}. First, 3 relevant documents $D_k$ for each AllenNLP-identified entity are retrieved using WikiAPI. Then, a BERT model trained on Wizard of Wikipedia (WoW) \cite{dinan:19} knowledge-response pairs ranks the top-10 evidence sentences $S_e$ from $D_k$. Next, to distinguish whether a response is making a factual claim, the lexical overlap between the response and $S_e$ is calculated, optimized on ABC-Eval training conversations. Finally, a RoBERTa classifier trained on Faith-Critic, a dataset of human-annotated faithful and unfaithful evidence-response pairs derived from WoW \cite{dziri_faithdial:22}, is applied to those responses that make a factual claim. If the response is predicted to be unfaithful to any evidence $e \in S_e$, then it is labelled as \texttt{!Fac}. 

\paragraph{S2T2}  

S2T2 is a semi-supervised student-teacher training framework using two teachers, one trained on the gold data and the other trained on perturbed gold data under a [MASK] replacement, to incorporate self-supervised data augmentation into the model training \cite{lin:22}. 
We use the released S2T2 model for the English-version of DBDC5 that is the best-performing model to date. 
We use S2T2 as identifying \texttt{Ign} and \texttt{!Rel} labels, since it is not trained to distinguish between them.

%S2T2 is a semi-supervised student-teacher training framework that utilizes two teachers - one trained on the gold data and one trained on perturbed gold data under a [MASK] replacement - to incorporate a self-trained data augmentation into the model training \cite{lin:22}. We use the released version of S2T2 for the English-version of DBDC5, which is the best-performing model to date. We apply S2T2 to \texttt{Ign} and \texttt{!Rel} labels, as it is not trained to distinguish between these labels.

\paragraph{DECODE (DEC)} 

We use the released RoBERTa classification model trained on DECODE to label \texttt{!Sel}. DECODE contains human-written contradictory and non-contradictory dialogue responses with respect to the current speaker's previous utterances in the dialogue \cite{nie:21}.

%\citet{nie:21} train a RoBERTa classification model on DECODE, a dataset with human-written contradictory and non-contradictory dialogue responses with respect to the current speaker's previous utterances in the dialogue. We use their released model to label \texttt{!Sel}.

% \footnote{\href{https://huggingface.co/ynie/roberta-large_conv_contradiction_detector_v0}{roberta-large\_conv\_contradiction\_detector\_v0(Huggingface)}}

\paragraph{EPITOME (EPI)} 

A RoBERTa-based bi-encoder classification model for each empathetic communication mechanism is trained from the publicly available Reddit portion of the EPITOME dataset \cite{sharma:20}. Predictions of weak or strong expressions of any of the three mechanisms are considered as \texttt{Emp}. Predictions of no expression for all mechanisms are considered as \texttt{!Emp}.

\section{LLM-based Behavior Detection}
\label{sec:gpt_llm}

For LLM-based dialogue behavior detection, we use OpenAI’s \textit{gpt-turbo-3.5-301} (henceforth, \LN ChatGPT).
Similar to the specialized models (Section~\ref{sec:models}), ChatGPT is tasked with classifying a single behavior at a time.
% The rest of this section details the prompt development process and the best-performing prompts.
Following the human annotator training process for ABC-Eval, we use the three training conversations for each label as our prompt engineering testbed.
%, curated to include the prevalent behaviors cases by \citet{finch:23}.
This section highlights key decisions of our prompt engineering process.

%Following the human annotator training process utilized by ABC-Eval, we take the 3 training conversations for each label as our testbed for ChatGPT prompt development, since these training conversations were specifically curated to consist of prevalent cases of the behaviors by \citet{finch:23}. In this section we highlight the major decisions made during the prompt development process.

\paragraph{Instruction Finetuning}

During prompt engineering, it became apparent that the instructions designed for human annotators (Section~\ref{sec:dataset}) were not suitable as ChatGPT instructions. 
We iteratively refined the instructions such that ChatGPT's mistakes on the training conversations were reduced. This involved removing instructions ChatGPT appeared to misunderstand as well as adding additional behavior details and specifications.

%Through testing, it became clear that the instructions optimized for human annotators provided by \citet{finch:23} were far from optimal for ChatGPT. Since ChatGPT appeared to be ignorant of much of the unspecified details of the behaviors, we iteratively updated the instructions to be more exhaustive and explicit towards defining each behavior.

\paragraph{Utterance Focus} 

We discovered that when ChatGPT was instructed to label each bot turn given the entire dialogue, the resulting classifications often focused on only a subset of the bot responses.
To ensure consistent and robust labeling for every bot utterance, our final prompt provides the dialogue history paired with the next bot response as the target utterance to be labeled.

%When the model was instructed to label each turn by Speaker 2 given an entire dialogue at once, we found that classifications were often considered on only a subset of the utterances. In order to guarantee control over obtaining a decision for each utterance, the final procedure provides a dialogue history paired with the next response as the target utterance to be labelled.

\paragraph{In-context Examples}

We also tried including the examples provided to human annotators by \citet{finch:23} as in-context examples in the prompts.
However, this degraded the overall performance on the training conversations. 
It appears that the examples optimized for improving \textit{human} annotations do not translate well to ChatGPT's performance.

%We also tried including the examples provided to human annotators by \citet{finch:23} as in-context examples in the prompts. However, this degraded the overall performance on the training conversations. Perhaps such examples optimized to improve \textit{human} annotations do not translate to ChatGPT.

\paragraph{Creativity}

We conducted experiments involving several \textit{temperature} parameters and observed high instability in the classifications for the same inputs when the temperature was increased. 
Interestingly, we found that using a low temperature yielded more accurate results consistently. 
Thus, we opted to use a \textit{temperature} of 0 for maximum reproducibility in our classifications. 
Similar findings have been reported by \citet{gilardi:23} and \citet{reiss:23}.

%We experimented with different \texttt{temperature} parameters and found high instability in the classifications for the same inputs as temperature increases. Often a low temperature produced more accurate results, thus we use a \texttt{temperature} of 0 for highest reproducibility. 
%Similar findings are reported by \citet{gilardi:23} and \citet{reiss:23}.

\paragraph{Final Prompt}

Table~\ref{tab:final_prompt} illustrates an example of the final prompt, in particular for the \texttt{Red} behavior\footnote{Due to spacing constraints, Table \ref{tab:final_prompt} contains minor discrepancies with the actual prompts in this work (Appendix \ref{app:full_prompt_example}).}. 
Each behavior is accompanied by its own eliciting question and description, which respectively fill in the \textit{Behavior Question} (\textbf{Q}) and \textit{Behavior Definition} (\textbf{D}) containers of the prompt. 
For labeling a particular context-response example, the historical turns from the context and the next target response fill in the \textit{Dialogue Context} (\textbf{C}) and \textit{Target Utterance} (\textbf{U}) containers.
The final label for a behavior is parsed from the line produced by ChatGPT that begins with the header `\textit{Decision:}', where the value `yes' indicates a positive label and any other value indicates a negative label.
Appendix \ref{app:prompt_values} provides the questions and definitions used for all behaviors.

\begin{table}[htbp!]
\centering\resizebox{\columnwidth}{!}{
\begin{tabular}{>{\columncolor[gray]{0.85}}p{0.1cm}l} 
\toprule
 & S1: Hi! \\
 & S2: Hi! I am happy to be talking to you. \\
 & $\quad\;\,$ Do you have any pets at home? \\
 & S1: I don't, my mom is allergic to most pets. \\
 & $\quad\;\,$ I used to have a pet fish when I was much younger. \\
 & S2: Do you want a pet at some point in the future? \\
 & $\quad\;\,$ No judgment here, I'm just curious. \\
 & S1: I would love to have pets in the future! Once I \\
\multirow{-9}{*}{\bf C}
 & $\quad\;\,$ have my own place, I'd like to get a dog or two. \\
\midrule
\bf Q & Is this response repeating something that has already been said: \\
\midrule
\bf U & S2: Would you want to get a cat or a dog? \\
\midrule
 & A response is repetitive if:\\
 & - it repeats something from earlier in the dialogue \\
 & - it includes asking a question whose answer has been \\
 & $\:\,$ already shared \\
 & If any part of the response is repetitive, then it should be \\
 & labeled as repetitive. \\
 & Note that sometimes repetition is useful, such as for emphasis, \\
 & acknowledgement, clarification, or elaboration, and in these \\
\multirow{-9}{*}{\bf D} 
 & cases it should NOT be labeled as repetitive. \\

\midrule
\multicolumn{2}{l}{Provide your reasoning when considering this question start-}\\
\multicolumn{2}{l}{ing with ``Reasoning:''. Then, finish by writing your final}\\
\multicolumn{2}{l}{decision as one of: ``Decision: [YES]'' or ``Decision: [NO]''.}\\
%\multicolumn{2}{l}{Do NOT fill in your decision with any terms other than YES or NO.} \\
\bottomrule
\end{tabular}}
\caption{A ChatGPT prompt example for the \texttt{Red} behavior. Segments in the prompt are dynamically modified based on the example and behavior, as highlighted in the gray containers (\textbf{C}: dialogue context, \textbf{Q}: behavior question, \textbf{U}: target utterance, \textbf{D}: behavior definition).}
\label{tab:final_prompt}
\vspace{-1.1em}
\end{table}

%\begin{figure}[hp!]
%    \centering
%    \includegraphics[width=\columnwidth]{latex/img/prompt.pdf}
%    \caption{A ChatGPT prompt example for the \texttt{Red} behavior. Segments in the prompt are dynamically modified based on the example and behavior, as highlighted in the gray containers (\textbf{C}: dialogue context, \textbf{Q}: behavior question, \textbf{U}: target next response, \textbf{D}: behavior definition).}
%    \label{fig:final_prompt}
%\vspace{-1em}    
%\end{figure}

%An example of the final prompt is shown in Figure \ref{fig:final_prompt} for the \texttt{Red} behavior. Each behavior receives its own eliciting question and textual description, which fill in the \textit{Behavior Question} (\textbf{Q}) and \textit{Behavior Definition} (\textbf{D}) containers of the prompt. 
%For labelling a particular context-response example, the historical turns from the context and target next response fill in the \textit{Dialogue Context} (\textbf{C}) and \textit{Target Utterance} (\textbf{U}) containers of the prompt. 
%Figure \ref{fig:gpt_prompts} in Appendix \ref{app:prompt_values} provides the complete set of questions and definitions used for the behaviors in this work. 
%The final label for a behavior is parsed from the line outputted by ChatGPT that begins with "Decision:", where a value of "yes" indicates a positive label and a value of "no" indicates a negative label.

%% file: latex/evaluation.tex
\section{Evaluation}
\label{sec:evaluation_metrics}

To evaluate the detection capability of the models in Sections~\ref{sec:models} and \ref{sec:gpt_llm}, we compare their performance against that of human annotators.
For this, we take the set of doubly annotated conversations in ABC-Eval as our evaluation set (108 dialogues), and apply each model to the bot responses (1,634 utterances) to obtain the predicted labels.

%To evaluate the detection capability of the models, we seek to compare their performance on the labelling tasks against those of human annotators. To this end, we take all of the doubly annotated conversations from \citet{finch:23} ($n=100$) as the evaluation dataset and apply each model to their utterances ($n=1634$) in order to obtain label results. By using the doubly annotated conversations, we obtain results for the automated methods that can be directly compared to the human performance.

% The identification of dialogue behaviors and errors is a challenging task, mainly because it is subjective and often up to human interpretation to what degree a particular response fits the behavior label, making it challenging to define a single label that can be considered the ultimate ground truth that all humans would agree on. Evaluations of these subjective dialogue characteristics are often aggregated over many human annotators in order to more robustly represent the general human consensus.

\subsection{Metrics}

To assess the degree to which automated methods can approximate human judgment for a particular dialogue behavior, we measure the accuracy of the binary labels predicted by automated methods with respect to the binary labels provided by the human annotators. In addition, we calculate both the F1-score for the positive occurrences of each dialogue behavior and for the negative occurrences of each dialogue behavior, in order to obtain a more fine-grained picture of the performance.

Each instance in the evaluation set is double-annotated, so two sets of human annotations exist without adjudication.
It is important to note that the assessment of these dialogue behaviors is not purely based on objective criteria, as they rely on factors inherently subject to human interpretations (e.g., commonsense contradiction, irrelevance).
With this in mind, to better capture the aggregate nature of identifying dialogue behaviors, the final score for each metric is measured by averaging results across the double human annotations, where $e$ is the metric (either accuracy or F1-score), $o_m$ is the model outputs, and $o_{h1}$ and $o_{h2}$ are the human labels from annotation round 1 and 2, respectively:
$$
    e_{final} = \frac{1}{2}(e(o_m, o_{h1}) + e(o_m, o_{h2}))
$$
To assess human performance, we measure the F1 score and accuracy by comparing the two human annotation sets.
Finally, the statistical significance between outputs of models and humans, and between outputs of the specialized models and ChatGPT, is estimated using McNemar's Test with significance level of $0.05$. Testing is performed by treating each human annotation set as ground-truth.\footnote{The other human annotation set relative to the one being treated as ground-truth is used as human output.}
% Finally, each pair of models is tested to determine whether their disagreement on each of the human annotation sets is statistically significant using McNemar's Test.

%To measure the extent to which an automated method is able to approximate a human judge, we calculate precision, recall, and F1-score on the human-labelled positive/negative instances (\textbf{(+/-) P/R/F1}) and accuracy over all instances (\textbf{Acc}). We also calculate human performance (\textbf{HUM}) by treating each round of annotations as the ground truth to benchmark the other annotation round against.

%Many of the dialogue behaviors are not purely objective phenomenon as they rely on criteria that are inherently up to interpretation by the human judge (e.g. commonsense contradition, irrelevant, etc.). 
%To better model the aggregative nature of identifying dialogue behaviors, we average the evaluation metrics across the double annotations provided by \citet{finch:23} such that:
%Even though there is not necessarily an obvious ground truth that all humans would agree on, \citet{finch:23} reports human agreement from 0.4 to 0.8 Krippendorff's alpha between their human annotators, indicating humans are able to achieve moderate agreement overall. 

\vspace{-0.5em}
\subsection{Results \& Discussion}
\label{sec:results}

\begin{table}[ht!]
    \centering
    {\small
    % \resizebox{\columnwidth}{!}{%
    \begin{tabular}{r|c|c|c|l|r}
    \toprule
    & \bf Model & \multicolumn{1}{|c}{\textbf{F1+}}& \multicolumn{1}{|c}{\textbf{F1-}} & \multicolumn{1}{|c}{\bf Acc.} & \multicolumn{1}{|c}{\bf \#+} \\
    \midrule
    \multirow{3}{*}{\tt Emp} & EPI & 54.2 & 31.3 & 45.0 & 1,343 \\
    & ChatGPT & 19.3 & 75.4 & \bf 62.3$^{\dagger\dagger}$ & 146 \\
     & HUM & 69.7 & 81.6 & \bf 77.1$^{\star\star}$ & 618 \\
    \midrule
    \multirow{3}{*}{\tt !Emp} & EPI & 13.4 & 83.5 & 72.3 & 291 \\
    & ChatGPT & 26.6 & 82.6 & 71.8 & 396 \\
    & HUM & 51.5 & 92.0 & \bf 86.3$^{\star\star}$ & 231 \\
    \midrule
    \multirow{2}{*}{\tt !Com} & ChatGPT & 34.9 & 86.7 & 78.0 & 219 \\
    & HUM & 55.6 & 88.6 & \bf 81.9$^{\star}$ & 333 \\
    \midrule
    \multirow{3}{*}{\tt !Fac} & FC & 15.9 & 90.1 & 82.2 & 223 \\
    & ChatGPT & 41.0 & 94.7 & \bf 90.3$^{\dagger\dagger}$ & 146 \\
    & HUM & 67.8 & 97.4 & \bf 95.2$^{\star\star}$ & 122 \\
    \midrule
    \multirow{3}{*}{\tt !Sel} & DEC & 31.1 & 92.6 & \bf 86.6$^{\dagger\dagger}$ & 215 \\
    & ChatGPT & 20.7 & 90.5 & 83.0 & 250 \\
    & HUM & 44.3 & 96.3 & \bf 93.1$^{\star\star}$ & 101 \\
    \midrule
    \multirow{2}{*}{\tt !Par} & ChatGPT & 18.6 & 93.8 & 88.5 & 79 \\
     & HUM & 48.8 & 94.8 & \bf 90.5$^{\star\star}$ & 151 \\
    \midrule
    \multirow{2}{*}{\tt Red} & ChatGPT & 32.9 & 93.8 & 88.6 & 148 \\
    & HUM & 58.7 & 96.4 & \bf 93.5$^{\star\star}$ & 129 \\
    \midrule
    \multirow{3}{*}{\tt Ign} &S2T2 & 25.2 & 85.3 & \bf 75.5$^{\dagger\dagger}$ & 365  \\
    & ChatGPT & 24.9 & 72.9 & 60.2 & 696 \\
     & HUM & 61.6 & 95.5 & \bf 92.0$^{\star\star}$ & 170 \\
    \midrule
    \multirow{3}{*}{\tt !Rel} & S2T2 & 27.9 & 82.9 & \bf 72.4$^{\dagger}$ & 365 \\
     & ChatGPT & 40.6 & 80.6 & 70.8 & 543 \\
     &HUM & 54.3 & 91.3 & \bf 85.4$^{\star\star}$ & 261 \\
    \bottomrule
    \end{tabular}
    }
    \caption{F1 and accuracy achieved by each model, where \textbf{HUM} stands for human judges. \textbf{\#+}: num. positive labels predicted. {\bm{$\dagger|\dagger\dagger$}} denote significance between \textit{automated} models on one or both human annotation sets, respectively. $\bm{\star|\star\star}$ denote significance against best automated model on one or both human annotation sets.}
    \label{tab:results}
\end{table}

\begin{table*}[ht!]
\centering\resizebox{\textwidth}{!}{
\begin{tabular}{c|c|l|r|c}
\toprule
\bf  Abbr.  & \bf Error Type & \multicolumn{1}{c|}{\bf Description} & \multicolumn{1}{c|}{$\bm{\Sigma}$} & \bf \%\\
\midrule    
\texttt{\textbf{IN}} & Inexperience          & Displays a lack of wisdom about human experiences & 83 & 0.23 \\
\texttt{\textbf{HF}} & History Forgetfulness & Forgets information shared previously in the history & 51 & 0.14 \\
\texttt{\textbf{DM}} & Definition Mismatch   & Expands beyond the provided definition of the behavior & 51 & 0.14 \\
\texttt{\textbf{SA}} & Selective Attention   & Overlooks components in a multi-idea response & 33 & 0.09 \\
\texttt{\textbf{DC}} & Disassociated Context & Incorrectly remembers the historical order of the conversation & 28 & 0.08 \\
\texttt{\textbf{SR}} & Semantic Relatedness  & Misunderstands the degree of similarity between two ideas & 19 & 0.05 \\
\texttt{\textbf{CN}} & Conversation Norms    & Misunderstands what constitutes a coherent progression of dialogue & 17 & 0.05 \\
\texttt{\textbf{ME}} & Mutual Exclusion      & Misidentifies when two events or concepts can or cannot co-occur together & 13 & 0.04 \\
\texttt{\textbf{RC}} & Role Confusion        & Confuses the speaker of previous utterances & 13 & 0.04 \\
\texttt{\textbf{MI}} & Misidentification     & Misunderstands the intent of what has been shared & 13 & 0.04 \\
\texttt{\textbf{CF}} & Confused Target       & Confuses which utterance is being labeled & 9 & 0.03 \\
\texttt{\textbf{TF}} & Temporal Framing      & Confuses the specified timeline of a particular situtation & 7  & 0.02 \\
\texttt{\textbf{RM}} & Reasoning Mismatch    & Its explanation is at-odds with its final decision & 7  & 0.02 \\
\texttt{\textbf{EX}} & Exhaustive            & Assumes all examples provided in the behavior definition must be met & 6 & 0.02 \\
\texttt{\textbf{CD}} & Claim Detection       & Incorrectly identifies when a claim/statement is being made & 4  & 0.01 \\
\texttt{\textbf{OA}} & Over-analysis         & Combines unrelated previous utterances to draw unsupported conclusions & 4  & 0.01 \\
\texttt{\textbf{BI}} & Bot Identity          & Considers indicators of speaker being a bot as erroneous & 2  & 0.01 \\
\bottomrule
\end{tabular}}
\caption{Results of the error analysis on ChatGPT's reasoning for dialogue behavior detection.}
\label{tab:error_analysis}
\vspace{-1em}
\end{table*}

\noindent Table~\ref{tab:results} indicates the ongoing challenge of dialogue behavior detection for automated models.
Across all labels, human judges are significantly more stable than the models.
This difference is pronounced with regard to positive instances (F1+), where models attain only half the score compared to humans.

Interestingly, ChatGPT exhibits comparable performance with several specialized classifiers.
In the case of \texttt{!Fac}, ChatGPT outperforms FaithCritic (FC) in every aspect and achieves performance closer to humans.
For \texttt{!Emp} and \texttt{!Rel}, ChatGPT shows similar performance on F1- and accuracy, and even better performance on F1+, as their classifiers.
Considering that ChatGPT is not finetuned for these tasks, these results are highly encouraging.

Although ChatGPT is seemingly outperformed by S2T2 on \texttt{Ign}, this is primarily due to the prediction of negative cases. When analyzing the positive cases, ChatGPT gives much higher recall yet similar precision compared to S2T2\footnote{Precision and recall provided in Appendix \ref{app:full_results}.}.
In practice, positive case detection is more impactful, implying that ChatGPT has an advantage in real-world applications.

Furthermore, although ChatGPT faces significant challenges in detecting positive cases of \texttt{Emp}, EPITOME (EPI) does not perform much better. Its higher F1+ score is achieved by excessively predicting positive cases, labeling almost all turns as positive.
This overprediction impairs its overall performance, allowing ChatGPT to outperform it when considering all cases as reflected in accuracy.

The only behavior for which ChatGPT appears to be beaten by the specialized classifier is against DECODE (DEC) for \texttt{!Sel}. However, the difference in performance is only slight overall. 

Notably, ChatGPT shows promising accuracy and negative F1 (F1-) to humans for the three behaviors for which specialized models are not available: \texttt{!Com}, \texttt{!Par}, and \texttt{Red}.
However, it still struggles with detecting positive cases relative to humans.

%% file: latex/error_analysis.tex
\section{ChatGPT Error Analysis}
\label{sec:gpt_error_analysis}

We perform an error analysis of ChatGPT's predictions of dialogue behaviors to better understand its limitations. 
For each dialogue behavior, we select 40 instances where ChatGPT and humans disagree, and examine the reasoning provided by ChatGPT
prior to its final decision (\textbf{\texttt{[R]}}; see examples below).
Table~\ref{tab:error_analysis} presents a set of dialogue characteristics and ChatGPT predispositions that highlight common mistakes made by ChatGPT across multiple dialogue behaviors.

%We perform an error analysis of ChatGPT's predictions of dialogue behaviors to gain insights into its current limitations. 

%We conduct an error analysis of the dialogue behavior predictions made by ChatGPT in order to better understand its current limitations. For each dialogue behavior, we take 40 instances of ChatGPT-to-human disagreement and examine the reasoning outputted by the ChatGPT model before it made its final decision (\res{in examples}). We construct a set of dialogue characteristics and ChatGPT predispositions in Table \ref{tab:error_analysis} that capture the mistakes of ChatGPT common across many of the dialogue behaviors.

% \footnote{There is no guarantee that the reasoning outputted by ChatGPT directly reflects its internal decision-making, but chain-of-thought prompting has been shown to be impactful on ChatGPT performance, so the reasoning outputted based on the prompts in this work is likely to at least be partially influential on the final decision}

%%%%%%%%%%%%%%%%%%%%%%
%
% Context Management
%
%%%%%%%%%%%%%%%%%%%%%%

\subsection{Context Management}

The predominant cause of ChatGPT's errors is its inability to focus on the pertinent parts of the dialogue and response.
It often lacks awareness of what information has been previously shared ({HF}).
In Figure~\ref{fig:history-forgetfulness}, the earlier response that already answers ``\textit{Are you older than your brother}'' is missed.

%The predominant cause of ChatGPT's errors is its inability to attend to the most relevant segments of the specified dialogue and response. It often displays a lack of awareness of what information has been previously shared (\textbf{HF}) as in the following example where the answer to the question "Are you older than your brother" was answered earlier:

\noindent Moreover, ChatGPT tends to disregard the immediately preceding turn ({DC}) and make its decision based on an older state of the dialogue.
In Figure~\ref{fig:disassociated-context}, it overlooks the preceding question about favorite colors.
Similarly, ChatGPT often misses ideas shared within a response, instead latching on the very last part of the dialogue ({SA}).
In Figure~\ref{fig:selective-attention}, ChatGPT fails to notice the emotional mimicry conveyed by S2 in response to S1's opinion about Nicki Minaj.

\begin{figure}[htp!]
\centering
\begin{tcolorbox}[colback=white,colframe=mgray,leftrule=2mm]
\begin{quote}
\small
\hum{I do! I have a younger brother. What about you?}

\sys{You have a brother? Cool! Do you guys get along?}

\hum{We used to fight a lot as kids, but we get along well now.}

\sys{That's awesome. Personally, I always wanted an older sister so I could have someone to look up to. Are you older than your brother?}

\vspace{-1em}
\hrulefill

\res{The response asks a new question that \textbf{has not been answered yet}, which is whether S1 is older than their brother. \textbf{This information has not been shared before}, so the response is not repetitive.}
\end{quote}
\end{tcolorbox}
\vspace{-0.5em}\caption{An example of History Forgetfulness (HF).}
\label{fig:history-forgetfulness}
\vspace{-0.6em}
\end{figure}

%\noindent ChatGPT is also vulnerable to ignoring the immediately previous turn of dialogue (\textbf{DC}), thus basing its decision on a previous state the dialogue was in. For example, it completely overlooks the previous question about favorite colors in Figure~\ref{fig:disassociated-context}.

%Similarly, it frequently misses ideas shared in a response, often latching on to the final thing that is said (\textbf{SA}). In this example, ChatGPT fails to notice the emotional mimicry that occurs from S2 sharing their opinion about Nicki Minaj in the response:

\vspace{-0,3em}
\begin{figure}[htp!]
\centering
\begin{tcolorbox}[colback=white,colframe=mgray,leftrule=2mm]
\begin{quote}
\small
\sys{I'm sure you'll feel great after your workout! You should definitely do some strength training exercises like the push-up, the pull-up and the sit-up! They're some of the most common bodyweight exercises!}

\hum{Yes! What's your favorite color?}

\sys{I'm not sure about you, but my favorite color is blue! I like it because it reminds me of the sea color of the sea!}

\vspace{-1em}
\hrulefill

\res{S1 asked about dinner, and S2 responded by saying they haven’t had dinner yet and then shifted the topic to working out. S2 then asked about S1’s favorite exercises at the gym. \textbf{S1 did not mention anything about favorite colors}, so S2’s response is completely off-topic and does not address the previous turn from S1.}
\end{quote}
\end{tcolorbox}
\vspace{-0.5em}\caption{An example of Disassociated Context (DC).}
\label{fig:disassociated-context}
\vspace{-1.2em}
\end{figure}

\begin{figure}[htp!]
\centering
\begin{tcolorbox}[colback=white,colframe=mgray,leftrule=2mm]
\begin{quote}
\small 
\sys{I love pop! What is your favorite R\&B artist? Do you like Nicki Minaj?}

\hum{I do like Nicki!}

\sys{I like her too, she's so talented. Do either of your parents work? My mom is a teacher.}

\vspace{-1em}
\hrulefill

\res{S2 does not demonstrate any understanding of S1’s emotions or react with an appropriate sentiment or emotion to S1’s shared experience. \textbf{Instead, S2 changes the subject to ask about S1’s parents}. This response is not empathetic.}
\end{quote}
\end{tcolorbox}
\vspace{-0.5em}\caption{An example of Selective Attention (SA).}
\label{fig:selective-attention}
\vspace{-0.6em}
\end{figure}

% \noindent ChatGPT also has a tendency to misrepresent previously shared information (\textbf{MI}). In the following example, S1's mention of pizza is not recognized by ChatGPT:

% \begin{tcolorbox}[colback=white,colframe=mgray,leftrule=2mm]
% \begin{quote}
% \small
% \hum{I like pasta and pizza.

% \sys{Pizza is great. I really like any kind of pizza. Do you have any pets?

% -------------------------------------------------------------------
% \res{S2's response completely ignores \textbf{S1's last turn about liking pasta}. Instead, S2 changes the topic to pizza and asks about pets. This response is off-topic and does not address S1's last turn.
% \end{quote}
% \end{tcolorbox}

\noindent In other cases, ChatGPT misattributes shared information to a wrong speaker ({RC}).
In Figure~\ref{fig:role-confusion}, it recalls that S2 indicated a return to normalcy, when in fact, it was S1 who made that claim.

%In other cases, ChatGPT misattributes shared information to a wrong speaker (\textbf{RC}), such as when it recalls that S2 indicated a return to normalcy when that claim belongs to S1 in actuality:

\begin{figure}[htp!]
\centering
\begin{tcolorbox}[colback=white,colframe=mgray,leftrule=2mm]
\begin{quote}
\small
\sys{Sounds like a good time!, so the corona virus has really changed the world, huh? What is it like where you live? Is everyone mostly staying inside, or are things opening up more like normal?}

\hum{Things are beginning to open up normal where I am but it has been a long 2 years. How about for you?}

\sys{I see., how long do you think it will take before the pandemic is totally over, and things go back to how they were before the virus?}

\vspace{-1em}
\hrulefill

\res{\textbf{S2's previous statement was about things opening up to normal where they are}, implying that the pandemic is not over yet. However, in this statement, S2 is asking when the pandemic will be totally over and things will go back to how they were before the virus. This seems to be a contradiction of previous statement.}
\end{quote}
\end{tcolorbox}
\vspace{-0.6em}\caption{An example of Role Confusion (RC).}
\label{fig:role-confusion}
\vspace{-1.2em}
\end{figure}

%%%%%%%%%%%%%%%%%%%%%%
%
% INSTRUCTION FOLLOWING
%
%%%%%%%%%%%%%%%%%%%%%%

\subsection{Instruction Following}

In addition, a frequent issue is that ChatGPT strays from the provided behavior definitions. In Figure \ref{fig:definition-mismatch}, esoteric knowledge is improperly considered as commonsense.
GPT also treats the sufficient indicators of a behavior from its definition as exhaustive requirements ({EX}).
In Figure~\ref{fig:exhaustive}, ChatGPT criticizes S2 for not indicating an emotion and offering support, despite S2's validation of S1's experience.

%Another common issue is that ChatGPT was vulnerable to straying from the behavior definitions. For instance, ChatGPT often over-generalized the definitions (\textbf{DM}), such as by encapsulating factual knowledge into commonsense:

%In addition, ChatGPT often treats the sufficient indicators of a particular behavior from its definition as exhaustive requirements (\textbf{EX}), like when it criticizes the following example for failing to indicate an emotion and offer reassurance, even though it acknowledges that the response does indeed show an understanding of S1's experience:

\vspace{-0.3em}
\begin{figure}[htp!]
\centering
\begin{tcolorbox}[colback=white,colframe=mgray,leftrule=2mm]
\begin{quote}
\small
\sys{What is the name of this dorm? Do you have a roommate? How are you liking it?}

\hum{It's MLAO. I don't have a roommate so it's great.}

\sys{Congratulations. Mlao is a great dorm then. It is the largest in chicago. Enjoy!}

\vspace{-1em}
\hrulefill

\res{The response would contradict commonsense because \textbf{MLAO is not the largest dorm in Chicago. In fact, there is no dorm in Chicago with that name.} This response makes an unfounded assumption and contradicts common knowledge.}
\end{quote}
\end{tcolorbox}
\vspace{-0.6em}
\caption{An example of Definition Mismatch (DM).}
\label{fig:definition-mismatch}
\vspace{-1.5em}
\end{figure}

\begin{figure}[ht!]
\centering
\begin{tcolorbox}[colback=white,colframe=mgray,leftrule=2mm]
\begin{quote}
\small
\sys{Cool! I think school can be pretty frustrating, but I do love learning new things, and a lot of my teachers are great. What about you? Do you like school?}

\hum{More or less. Doing activities like talking to a robot is great, but doing assignments are tired. I've been spending the past week doing homework.}

\sys{I understand, it can definitely be stressful. What would you change about your school, if you could?}

\vspace{-1em}
\hrulefill

\res{S2 acknowledges that school can be frustrating and asks what S1 would change about it, \textbf{showing an understanding of S1's experience}. However, \textbf{they do not react with a specific sentiment or emotion, nor do they offer reassurance or support}.}
\end{quote}
\end{tcolorbox}
\vspace{-0.9em}
\caption{An example of Exhaustive (EX).}
\label{fig:exhaustive}
\vspace{-0.5em}
\end{figure}

\subsection{World Model}

Another major issue is ChatGPT's poor understanding of common world events and human experiences.

ChatGPT frequently reveals a limited understanding of the relationship among concepts within a dialogue. 
For example, it overlooks elaborations on previous points, considering them too semantically similar (SR).
In Figure~\ref{fig:semantic-relatedness}, the opinion about the cat's characteristics is actually unique information relative to the context.

\begin{figure}[htp!]
\centering
\begin{tcolorbox}[colback=white,colframe=mgray,leftrule=2mm]
\begin{quote}
\small
\sys{I love animals too! My favorite animal is the cat! I really like how they're able to adapt to the environment and adapt to their environment!}

\hum{I love cats too, theya re adorable!}

\sys{I love how cute they are! I especially like how their fur is so soft and fluffy! They look so beautiful!}

\vspace{-1em}
\hrulefill

\res{The response mentions how cute cats are and how soft and fluffy their fur is. While this is related to S2's favorite animal being cats, \textbf{it does not add any new information to the conversation} and is repetitive. Therefore, the response is repetitive.}
\end{quote}
\end{tcolorbox}
\vspace{-0.5em}\caption{An example of Semantic Relatedness (SR).}
\label{fig:semantic-relatedness}
\vspace{-0.5em}
\end{figure}

\noindent ChatGPT also often criticizes typical conversational practices ({CN}). In Figure~\ref{fig:conversation-norms}, it considers topical introductions at the start of a dialogue as irrelevant.

%Another significant issue arises from ChatGPT having a poor understanding of common world events and human experiences. In Figure~\ref{fig:conversation-norms}, many typical conversational practices are criticized by ChatGPT (\textbf{CN}) like when it assigns topical introductions at the start of a dialogue as being irrelevant:

\begin{figure}[htp!]
\centering
\begin{tcolorbox}[colback=white,colframe=mgray,leftrule=2mm]
\begin{quote}
\small
\hum{Hi!}

\sys{hi! I am happy to be talking to you. I hope you don’t mind me asking, but are you dating anyone now?}

\vspace{-1em}
\hrulefill

\res{The response from S2 is completely irrelevant to what S1 said. S1 simply greeted S2 and \textbf{did not provide any information that would lead to a discussion about dating}. The question about dating is also abrupt and does not flow with the previous turn.}
\end{quote}
\end{tcolorbox}
\vspace{-0.5em}\caption{An example of Conversation Norms (CN).}
\label{fig:conversation-norms}
\vspace{-0.5em}
\end{figure}

\noindent In addition, ChatGPT faces challenges in comprehending the plausibility of co-occurring events, beliefs, and experiences ({ME}). 
In Figure~\ref{fig:mutual-exclusion}, it depicts a lack of understanding that it is implausible to have an opinion about the music of an artist if one has no prior experience with that artist's music.

%\noindent In addition, ChatGPT struggles with understanding the plausibility of co-occurring events, beliefs, and experiences ({ME}). In Figure~\ref{fig:mutual-exclusion}, it does not appear to understand that it is implausible to have an opinion about the music of an artist if you have no experience with that artist's music:

ChatGPT also demonstrates a large degree of general inexperience with common phenomenon or situations in the world, which can lead to it harboring untrue facts about the world or misunderstanding nuances of situations ({IN}). In Figure~\ref{fig:inexperience}, ChatGPT's lack of commonsense is highlighted by its acceptance of well-wishes for a non-existent trip.

\begin{figure}[ht]
\centering
\begin{tcolorbox}[colback=white,colframe=mgray,leftrule=2mm]
\begin{quote}
\small
\sys{I have not heard of doj cat. Where does she live? Have any albums yet?}

\hum{He most recent album is "Planet Her". You should definitely listen to it!}

\sys{Planet her is amazing..the whole album has a distinct sound. It was produced by alex pall.}

\vspace{-1em}
\hrulefill

\res{There is no indication in the previous dialogue that would suggest a self-contradiction in S2's statement. \textbf{They are simply expressing their positive opinion about the album "Planet Her" and mentioning the producer}. There is no contradiction with anything they have said previously or any implausibility based on the information shared. Additionally, there is no contradiction with anything S1 has said. Therefore, there is no self-contradiction in S2's statement.}
\end{quote}
\end{tcolorbox}
\vspace{-0.5em}\caption{An example of Mutual Exclusion (ME).}
\label{fig:mutual-exclusion}
\vspace{-0.5em}
\end{figure}

\begin{figure}[htp!]
\centering
\begin{tcolorbox}[colback=white,colframe=mgray,leftrule=2mm]
\begin{quote}
\small
\hum{I don’t have a trip planned, but maybe I’ll go somewhere soon}

\sys{I’m sure you will! You’ll love it! Have a wonderful trip! I’ll see you soon!}

\vspace{-1em}
\hrulefill

\res{The response does not contradict commonsense as it is a polite and friendly response to the previous statement. It does not make any unfounded assumptions, is not illogical or self-contradictory, and does not ask a question where the answer is already obvious. \textbf{It is a common response to wish someone a good trip even if they have not planned one yet}.}
\end{quote}
\end{tcolorbox}
\vspace{-0.5em}\caption{An example of Inexperience (IN).}
\label{fig:inexperience}
\vspace{-1em}
\end{figure}

%% file: latex/recommendation.tex
\section{Recommendations}
\label{sec:recommendations}

Given the compelling performance for many dialogue behaviors observed in this work, ChatGPT is a promising direction for behavior classification.

For one, it is worth noting that ChatGPT boasts extreme cost-efficiency relative to humans. Where ChatGPT costs \$0.02 on average to provide labels for a single behavior for one dialogue in this work, the average cost for human annotation ranges from \$0.29 to \$1.96 depending on the behavior (Table~\ref{tab:cost_comparison} in Appendix~\ref{sec:appendix_cost}). Since even specialized classifiers rely on human annotations for training creation, they also end up being quite costly to maintain. 

Furthermore, the results of our error analysis reveal a large degree of systematicity behind ChatGPT's reasoning mistakes across many of the behaviors. Correcting these common mistakes is likely to further improve its performance to a noticeable degree. We next discuss mitigation strategies of these identified issues to aid in future work.

\paragraph{Context Management} Providing the complete dialogue history may hinder ChatGPT's ability to attend to the salient content due to information overload. To address this, we highlight two strategies:
\begin{itemize}
    \item \textit{Windowed Context}: instead of providing the entire history, truncate the context to \textit{k} previous turns. This would directly restrict the decision-making to the immediate context, which is important for behaviors that depend on accurate recency identification, including \texttt{!Rel}, \texttt{Ign}, \texttt{!Emp}, and \texttt{Emp}.
    \item \textit{Turn Pairing}: perform the labeling relative to each historical turn segment independently, rather than a contiguous context. This would enable explicit and focused comparisons to smaller segments of the history that could aid behaviors that require such precision, including \texttt{!Sel}, \texttt{!Par}, and \texttt{Red}.
\end{itemize}
% Such strategies could address the HF, DC, CT, and SA mistakes, which comprise 33\% of the identified mistakes.

\paragraph{In-Context Learning Examples} Given the identified mistake types, it becomes more straightforward to compose useful in-context learning examples that are tailored to optimizing ChatGPT. Examples of those mistake types that are related to ChatGPT misunderstanding the nuances of a behavior (e.g. MD, SR, CN, ME, EX) could be taken from a held-out set of conversations, which would prime ChatGPT to avoid such reasoning.

%% file: latex/limitations.tex
\section{Limitations}

Although ChatGPT is a high-performing, widely accessible, and affordable LLM at the time of writing, there are considerations towards the long-term applicability of the results found in this work due to the ChatGPT infrastructure. Since ChatGPT is not open-source and is only accessible through a paid API, there is less detailed understanding of its training and model design. In addition, this access method for ChatGPT also results in less user control over potential model changes and even model deprecation over time. As such, further studies could assess the applicability of other language models to the task of dialogue behavior detection to mitigate these concerns, and we leave this to future work.

Furthermore, it should be noted that the errors made by ChatGPT may not necessarily align with those made by alternative open-source language models, or even future versions of ChatGPT itself. However, it may still be useful to be mindful of the prominent problems encountered with ChatGPT while using other LLMs. These identified phenomena play a crucial role in language comprehension and reasoning overall and could also present challenges for other models, although the extent of their impact remains to be explored.

%However, the issues that are prominent for ChatGPT should be kept in mind when utilizing LLMs, regardless of the underlying model itself, since these identified phenomenon are often underpinnings of language understanding and reasoning in general and could be pitfalls for other models as well, although the extent remains to be seen. 

% for assessing current progress on the task of automated dialogue behavior detection at the time of writing,

%% file: latex/conclusion.tex
\section{Conclusion}
\label{sec:conclusion}

Although automated methods for dialogue behavior classification remain a challenging task, this work finds that ChatGPT-3.5 presents promising potential to reduce the gap between model and human performance. ChatGPT's ability to provide competitive behavior classification against specialized classifiers without necessitating finetuning or human annotation across a variety of dialogue behaviors gives rise to a low-cost, multi-task evaluator model. The systematicity behind the common mistakes observed for ChatGPT reveal concrete steps for future improvements that will improve behavior classification performance, including strategies for context management and better understanding of situational nuances. We look forward to future advancements in behavior classification that leverage ChatGPT's unique capabilities.

%% file: latex/acknowledgements.tex
\section{Acknowledgements}

We gratefully acknowledge the support of the Amazon Alexa AI grant. Any opinions, findings, and conclusions or recommendations expressed in this material are those of the authors and do not necessarily reflect the views of Amazon.

%% file: latex/appendix.tex
\section{Behavior Questions and Definitions}
\label{app:prompt_values}

The Question (\textbf{Q}) and Definition (\textbf{D}) for each dialogue behavior label used for the final ChatGPT prompts are shown in Tables \ref{tab:emp_prompt} - \ref{tab:!rel_prompt}, excluding \texttt{Red} which is shown in Table \ref{tab:final_prompt} in Section \ref{sec:gpt_llm}.

% \begin{figure*}[t]
% \centering
% \begin{subfigure}{0.85\textwidth}
%   \centering
%   \includegraphics[width=\linewidth]{latex/img/behavior_1.pdf}
% \end{subfigure}
% \begin{subfigure}{0.85\textwidth}
%   \centering
%   \includegraphics[width=\linewidth]{latex/img/behavior_2.pdf}
% \end{subfigure}
% \begin{subfigure}{0.85\textwidth}
%   \centering
%   \includegraphics[width=\linewidth]{latex/img/behavior_3.pdf}
% \end{subfigure}
% \caption{The Question \textbf{[Q]} and Definitions \textbf{[D]} supplied to the ChatGPT prompt for each dialogue behavior.}
% \label{fig:gpt_prompts}
% \end{figure*}

\begin{table}[htbp!]
\centering\resizebox{\columnwidth}{!}{
\begin{tabular}{>{\columncolor[gray]{0.85}}p{0.1cm}l} 
\toprule
\bf Q & Is this an empathetic response by Speaker 2: \\
\midrule
 & A response is empathetic when Speaker 2 does ONE \\
 & of the following: \\
 & - clearly demonstrates an understanding of Speaker 1's \\
 & $\:\,$ emotions \\
 & - reacts with the appropriate sentiment or emotion \\
 & $\:\,$ to Speaker 1's shared experience \\ 
\multirow{-5}{*}{\bf D} 
 & - understands or appropriately reacts to Speaker 1's \\
 & $\:\,$ experience or emotions \\
 & - appropriately reassures, encourages, or supports Speaker 1 \\
\bottomrule
\end{tabular}}
\caption{\texttt{Emp}: behavior question and definition.}
\label{tab:emp_prompt}
\vspace{-1em}
\end{table}

\begin{table}[htbp!]
\centering\resizebox{\columnwidth}{!}{
\begin{tabular}{>{\columncolor[gray]{0.85}}p{0.1cm}l} 
\toprule
\bf Q & If this were the next response in the dialogue, \\
      & would Speaker 1 feel like their feelings are not \\
      & being understood by Speaker 2: \\
\midrule
 & A response displays a lack of empathy when: \\
 & - it indicates a misunderstanding of how Speaker 1 \\
 & $\:\,$ feels based on what Speaker 1 just said \\
 & - the tone, emotion, or sentiment of the response is \\
 & $\:\,$ clearly inappropriate for what Speaker 1 just said \\
 & - the response has an inappropriate lack of emotion to \\
 & $\:\,$ what Speaker 1 just said \\
 & Do NOT consider its empathy relative to previous topics in \\
 & the conversation if the dialogue has moved on from them. \\
\multirow{-9}{*}{\bf D} 
 & Instead, only consider the most recent dialogue context \\
 & when evaluating the empathy of a response. \\
\bottomrule
\end{tabular}}
\caption{\texttt{!Emp}: behavior question and definition.}
\label{tab:!emp_prompt}
\vspace{-1em}
\end{table}

\begin{table}[htbp!]
\centering\resizebox{\columnwidth}{!}{
\begin{tabular}{>{\columncolor[gray]{0.85}}p{0.1cm}l} 
\toprule
 \bf Q & If this were the next response in the dialogue, \\
       & would it contradict commonsense: \\
\midrule
 & To identify contradictions of commonsense, judge whether \\
 & a vast majority of people would agree that the response \\
 & doesn't make sense because the response: \\
 & - contradicts common knowledge \\
 & - makes unfounded assumptions \\
 & - is highly illogical or self-contradictory \\
 & - asks a question where the answer is already obvious \\
 & Do NOT mark responses that don't make sense because they: \\
 & - are off-topic or irrelevant as responses \\
\multirow{-9}{*}{\bf D} 
 & - don't have any clear meaning (e.g. overly vague or  \\
 & $\:\,$ ill-formed responses)  \\
\bottomrule
\end{tabular}}
\caption{\texttt{!Com}: behavior question and definition.}
\label{tab:!com_prompt}
\vspace{-1em}
\end{table}

\begin{table}[htbp!]
\centering\resizebox{\columnwidth}{!}{
\begin{tabular}{>{\columncolor[gray]{0.85}}p{0.1cm}l} 
\toprule
\bf Q & If this were the next response in the dialogue, does it \\
      & completely ignore the immediate last turn from Speaker 1: \\
\midrule
 & Responses that are completely off-topic, fail to address the \\ 
 & asked question, or are otherwise completely inappropriate in \\
\multirow{-3}{*}{\bf D} 
 & the context are considered to be ignoring the other speaker. \\
\bottomrule
\end{tabular}}
\caption{\texttt{Ign}: behavior question and definition.}
\label{tab:ign_prompt}
\vspace{-1em}
\end{table}

\begin{table}[htbp!]
\centering\resizebox{\columnwidth}{!}{
\begin{tabular}{>{\columncolor[gray]{0.85}}p{0.1cm}l} 
\toprule
\bf Q & If this were the next response in the dialogue, \\
      & is it a self-contradiction by Speaker 2: \\
\midrule
 & Self contradictions occur when Speaker 2 says \\
 & something that is a contradiction of what they have \\
 & said previously or it is extremely implausible based \\
 & on the information they have already shared. \\
 & Self contradictions may also occur within a single turn \\
 & if Speaker 2 shares two contradictory things. \\
 & If Speaker 2 shares world knowledge that is factually \\
 & incorrect this is NOT enough on its own to warrant a \\
 & self contradiction. \\
\multirow{-9}{*}{\bf D} 
 & If Speaker 2 contradicts something the other speaker \\
 & Speaker 1 has said, this is NOT a self-contradiction. \\
\bottomrule
\end{tabular}}
\caption{\texttt{!Sel}: behavior question and definition.}
\label{tab:!sel_prompt}
\vspace{-1em}
\end{table}

\begin{table}[htbp!]
\centering\resizebox{\columnwidth}{!}{
\begin{tabular}{>{\columncolor[gray]{0.85}}p{0.1cm}l} 
\toprule
\bf Q & Does this response include an incorrect fact: \\
\midrule
 & Incorrect facts occur when the response includes \\
 & information that is either: \\
 & - false \\
 & - unproven \\
 & - highly controversial \\
 & - highly implausible \\
 & - clearly misleading \\
 & If an organization, person, place, etc. is mentioned as a \\
 & part of public knowledge, but it does not exist or it is  \\
 & inaccurately represented, then this is an incorrect fact. \\
 & Do NOT consider a turn as an incorrect fact if the turn could \\
 & be interpreted as expressing: \\
 \multirow{-9}{*}{\bf D} 
 & - preference or value judgements \\
 & - estimates or predictions \\
 & - personal information about the speaker or their partner \\
 & - information about things in either speaker's life that are \\
 & $\:\,$ not publicly relevant \\
\bottomrule
\end{tabular}}
\caption{\texttt{!Fac}: behavior question and definition.}
\label{tab:!fac_prompt}
\vspace{-1em}
\end{table}

\begin{table}[htbp!]
\centering\resizebox{\columnwidth}{!}{
\begin{tabular}{>{\columncolor[gray]{0.85}}p{0.1cm}l} 
\toprule
\bf Q & Is Speaker 2 saying something about Speaker 1 that \\
      & is contradicting what Speaker 1 has already shared: \\
\midrule
 & Partner contradictions occur when Speaker 2:\\
 & - shares an assumption about Speaker 1 that is impossible \\
 & $\:\,$ to know based on what has already been said \\
 & - shares an inference about Speaker 1 that is implausible \\
 & $\:\,$ based on what has already been said \\
 & - contradicts something Speaker 1 shared about themselves \\
 & - asks a repetitive question about Speaker 1 when the \\
 & $\:\,$ answer is already known based on what has already been said \\
 & If Speaker 2 says something that makes it seem like they have \\
 & forgotten or misremembered what their partner Speaker 1 \\
 & has said earlier in the dialogue, this is a partner contradiction. \\
\multirow{-10}{*}{\bf D} 
 & If Speaker 2 shares a difference of opinion or \\
 & situation in their own life as compared to Speaker 1, \\
 & this is NOT a partner contradiction. \\
\bottomrule
\end{tabular}}
\caption{\texttt{!Par}: behavior question and definition.}
\label{tab:!par_prompt}
\vspace{-1em}
\end{table}

\begin{table}[htbp!]
\centering\resizebox{\columnwidth}{!}{
\begin{tabular}{>{\columncolor[gray]{0.85}}p{0.1cm}l} 
\toprule
\bf Q & If this were the next response in the dialogue, \\
      & is it completely irrelevant to what was just said: \\
\midrule
 & If a response fails to continue the current discussion or jumps to \\
 & a new and off-topic discussion, it is considered to be irrelevant. \\
 & Responses that are irrelevant feel abrupt and interrupt the \\
 & discussion, usually because they present questions or ideas \\
 & that are unrelated to the previous turn. Short reactions to or \\
 \multirow{-6}{*}{\bf D} 
 & acknowledgements of the previous turn are NOT irrelevant.  \\
\bottomrule
\end{tabular}}
\caption{\texttt{!Rel}: behavior question and definition.}
\label{tab:!rel_prompt}
\vspace{-1em}
\end{table}

% \begin{table}[htbp!]
% \centering\resizebox{\columnwidth}{!}{
% \begin{tabular}{>{\columncolor[gray]{0.85}}p{0.1cm}l} 
% \toprule
% \bf Q & Is this response repeating something that has already been said: \\
% \midrule
%  & A response is repetitive if:\\
%  & - it repeats something from earlier in the dialogue \\
%  & - it includes asking a question whose answer has been \\
%  & $\:\,$ already shared \\
%  & If any part of the response is repetitive, then it should be \\
%  & labeled as repetitive. \\
%  & Note that sometimes repetition is useful, such as for emphasis, \\
%  & acknowledgement, clarification, or elaboration, and in these \\
% \multirow{-9}{*}{\bf D} 
%  & cases it should NOT be labeled as repetitive. \\
% \bottomrule
% \end{tabular}}
% \caption{\texttt{Red}: behavior question and definition.}
% \label{tab:red_prompt}
% \vspace{-1em}
% \end{table}

\newpage

\section{Full Prompt Example}
\label{app:full_prompt_example}

Table \ref{tab:full_final_prompt} shows an example of the full ChatGPT prompt utilized in this work. This full version preserves the `DIALOGUE' header, turn numbers, whitespace newlines, and full speaker names, which were removed from the prompt in Table \ref{tab:final_prompt} in Section \ref{sec:gpt_llm} due to spacing constraints.

\begin{table}[htbp!]
\centering\resizebox{\columnwidth}{!}{
\frame{\begin{tabular}{l} 
- - - - - - - - - - \\
DIALOGUE \\
- - - - - - - - - - \\
1. Speaker 1: Hi! \\
2. Speaker 2: Hi! I am happy to be talking to you. Do you have any \\
$\:\,$$\:\,$ pets at home? \\
3. Speaker 1: I don't, my mom is allergic to most pets. I used to have \\
$\:\,$$\:\,$ a pet fish when I was much younger. \\
4. Speaker 2: Do you want a pet at some point in the future? No \\
$\:\,$$\:\,$ judgment here, I'm just curious. \\
5. Speaker 1: I would love to have pets in the future! Once I have \\
$\:\,$$\:\,$ my own place, I'd like to get a dog or two. \\
- - - - - - - - - - \\
\\
Is this response repeating something that has already been said: \\
\\
Speaker 2: Would you want to get a cat or a dog? \\
\\
A response is repetitive if:\\
- it repeats something from earlier in the dialogue \\
- it includes asking a question whose answer has been \\
$\:\,$ already shared \\
\\
If any part of the response is repetitive, then it should be \\
labeled as repetitive. \\
Note that sometimes repetition is useful, such as for emphasis, \\
acknowledgement, clarification, or elaboration, and in these \\
cases it should NOT be labeled as repetitive. \\
\\
Provide your reasoning when considering this question starting \\
with ``Reasoning:''. Then, finish by writing your final \\
decision as one of: ``Decision: [YES]'' or ``Decision: [NO]''. \\
Do NOT fill in your decision with any terms other than YES or NO. \\
\end{tabular}}}
\caption{An example of an unmodified ChatGPT prompt.}
\label{tab:full_final_prompt}
\vspace{-1.5em}
\end{table}

\section{Full Results}
\label{app:full_results}

Table \ref{tab:full_results} extends Table \ref{tab:results} from \S\ref{sec:results} to include the precision and recall scores for the automated models. Precision and recall scores are not meaningful for the human evaluators since each human annotation set is traded out as a benchmark against the other; thus, we still present only F1 for \textbf{HUM}.

\begin{table}[hb!]
    \centering
    {\small
    \resizebox{\columnwidth}{!}{%
    \begin{tabular}{r|c|r|r|l|l}
    \toprule
    & \bf Model & \multicolumn{1}{|c|}{\textbf{P/R/F1+}}& \multicolumn{1}{|c|}{\textbf{P/R/F1-}} & \multicolumn{1}{|c|}{\bf Acc.} & \multicolumn{1}{|c}{\bf \#$\bm{_+}$} \\
    \midrule
    \multirow{3}{*}{\tt !Fac} & FC & 12.3 / 22.4 / 15.9 & 93.3 / 87.1 / 90.1 & 82.2 & 223 \\
    & ChatGPT & 37.7 / 44.9 / 41.0 & 95.5 / 94.0 / 94.7 & \bf 90.3$^{\dagger\dagger}$ & 146 \\
    & HUM & 67.8 & 97.4 & \bf 95.2$^{\star\star}$ & 122 \\
    \hline %&&&&&& \\ 
    \multirow{2}{*}{\tt Red} & ChatGPT & 30.7 / 35.5 / 32.9 & 94.3 / 93.2 / 93.8 & 88.6 & 148 \\
    & HUM & 58.7 & 96.4 & \bf 93.5$^{\star\star}$ & 129 \\
    \hline %&&&&&& \\
    \multirow{2}{*}{\tt !Com} & ChatGPT & 43.8 / 29.1 / 34.9 & 83.3 / 90.5 / 86.7 & 78.0 & 219 \\
    & HUM & 55.6 & 88.6 & \bf 81.9$^{\star}$ & 333 \\
    \hline %&&&&&& \\
    \multirow{3}{*}{\tt !Rel} & S2T2 & 24.0 / 33.5 / 27.9 & 86.3 / 79.8 / 82.9 & \bf 72.4$^{\dagger}$ & 365 \\
     & ChatGPT & 30.1 / 62.5 / 40.6 & 91.0 / 72.3 / 80.6 & 70.8 & 543 \\
     & HUM & 54.3 & 91.3 & \bf 85.4$^{\star\star}$ & 261 \\
    \hline %&&&&&& \\
    \multirow{2}{*}{\tt !Par} & ChatGPT & 27.2 / 14.2 / 18.6 & 91.6 / 96.1 / 93.8 & 88.5 & 79 \\
     & HUM & 48.8 & 94.8 & \bf 90.5$^{\star\star}$ & 151 \\
    \hline %&&&&&& \\
    \multirow{3}{*}{\tt !Sel} & DEC & 22.8 / 49.1 / 31.1 & 96.3 / 89.2 / 92.6 & \bf 86.6$^{\dagger\dagger}$ & 215 \\
    & ChatGPT & 14.6 / 35.9 / 20.7 & 95.3 / 86.1 / 90.5 & 83.0 & 250 \\
    & HUM & 44.3 & 96.3 & \bf 93.1$^{\star\star}$ & 101 \\
    \hline %&&&&&& \\
    \multirow{3}{*}{\tt !Emp} & EPI & 12.0 / 15.1 / 13.4 & 85.4 / 81.8 / 83.5 & 72.3 & 291 \\
    & ChatGPT & 21.1 / 36.2 / 26.6 & 88.1 / 77.7 / 82.6 & 71.8 & 396 \\
    & HUM & 51.5 & 92.0 & \bf 86.3$^{\star\star}$ & 231 \\
    \hline %&&&&&& \\
    \multirow{3}{*}{\tt Ign} &S2T2 & 18.5 / 39.5 / 25.2 & 91.9 / 79.7 / 85.3 & \bf 75.5$^{\dagger\dagger}$ & 365  \\
    & ChatGPT & 15.5 / 63.4 / 24.9 & 93.3 / 59.8 / 72.9 & 60.2 & 696 \\
     & HUM & 61.6 & 95.5 & \bf 92.0$^{\star\star}$ & 170 \\
    \hline %&&&&&& \\
    \multirow{3}{*}{\tt Emp} & EPI & 39.6 / 86.0 / 54.2 & 70.3 / 20.1 / 31.3 & 45.0 & 1343 \\
    & ChatGPT & 50.7 / 11.9 / 19.3 & 63.4 / 92.9 / 75.4 & \bf 62.3$^{\dagger\dagger}$ & 146 \\
     & HUM & 69.7 & 81.6 & \bf 77.1$^{\star\star}$ & 618 \\
    \bottomrule
    \end{tabular}
    }
    }
    \caption{Precision, recall, F1 and accuracy achieved by each model, where \textbf{HUM} stands for human judges. \textbf{\#+}: num. positive labels predicted. {\bm{$\dagger|\dagger\dagger$}} denote significance between \textit{automated} models on one or both annotation sets. $\bm{\star|\star\star}$ denote significance against best automated model on one or both annotation sets, respectively.}
    \label{tab:full_results}
\end{table}

\section{ChatGPT Cost}
\label{sec:appendix_cost}

We compare the average cost of labeling a single dialogue from ABC-Eval for each behavior using ChatGPT and human judges. Table \ref{tab:cost_comparison} contains the calculated costs.

\paragraph{ChatGPT} The ChatGPT cost for a single dialogue is calculated from the OpenAI API pricing\footnote{\url{https://openai.com/pricing}} \LN (\$0.002 USD per 1000 tokens, at time of writing) on the sum total number of tokens used for obtaining labels for each bot response for a particular behavior. These costs are then averaged over all dialogues used in this work to obtain the average cost per dialogue. Because there is not much difference in prompt length for the different behavior prompts, the average cost per behavior is quite similar.

\paragraph{HUM} Human annotation costs are derived from the average costs presented in \citet{finch:23}. Since the behavior labels were grouped into annotation tasks for the human judges, we divide each task cost by the number of behaviors contained within that task. The cost for a single label is then the resulting quotient for its respective task. 

\input{latex/cost_table}

%% file: latex/cost_table.tex
\begin{table}[thb!]
\centering
\resizebox{.5\columnwidth}{!}{%
\begin{tabular}{c|cl}
\toprule
 & \bf ChatGPT & \bf HUM \\
\midrule
\texttt{!Fac} & 0.02 & 1.96 \\
\texttt{Red} & 0.02 & 0.29 \\
\texttt{!Com} & 0.02 & 0.92 \\
\texttt{!Rel} & 0.02 & 0.47 \\
\texttt{!Par} & 0.02 & 0.29 \\
\texttt{!Sel} & 0.02 & 0.29 \\
\texttt{!Emp} & 0.02 & 0.58 \\
\texttt{Ign} & 0.02 & 0.47 \\
\texttt{Emp} & 0.02 & 0.58 \\
\bottomrule
\end{tabular}}
\caption{Cost (\$ USD) per dialogue for each behavior using ChatGPT or humans (\textbf{HUM}).}
\label{tab:cost_comparison}
\end{table}